\documentclass{article}

\usepackage{PRIMEarxiv}

\usepackage[utf8]{inputenc} % allow utf-8 input
\usepackage[T1]{fontenc}    % use 8-bit T1 fonts
\usepackage{hyperref}       % hyperlinks
\usepackage{url}            % simple URL typesetting
\usepackage{booktabs}       % professional-quality tables
\usepackage{amsfonts}       % blackboard math symbols
\usepackage{nicefrac}       % compact symbols for 1/2, etc.
\usepackage{microtype}      % microtypography
\usepackage{lipsum}
\usepackage{fancyhdr}       % header
\usepackage{graphicx}       % graphics
\graphicspath{{media/}}     % organize your images and other figures under media/ folder

\title{Classification of assembly tasks combining multiple primitive actions using Transformers
and xLSTMs}
\author{
  Miguel Neves \\
  University of Coimbra \\
  Coimbra, Portugal\\
  \texttt{miguel.neves@dem.uc.pt} \\
   \And
  Pedro Neto \\
  University of Coimbra \\
  Coimbra, Portugal \\
  \texttt{pedro.neto@dem.uc.pt} \\
}

\begin{document}
\maketitle

\begin{abstract}
The classification of human-performed assembly tasks is essential in collaborative robotics to ensure safety, anticipate robot actions, and facilitate robot learning. However, achieving reliable classification is challenging when segmenting tasks into smaller primitive actions is unfeasible, requiring us to classify long assembly tasks that encompass multiple primitive actions. In this study, we propose classifying long assembly sequential tasks based on hand landmark coordinates and compare the performance of two well-established classifiers, LSTM and Transformer, as well as a recent model, xLSTM. We used the HRC scenario proposed in the CT benchmark, which includes long assembly tasks that combine actions such as insertions, screw fastenings, and snap fittings. Testing was conducted using sequences gathered from both the human operator who performed the training sequences and three new operators. The testing results of real-padded sequences for the LSTM, Transformer, and xLSTM models was 72.9\%, 95.0\% and 93.2\% for the training operator, and 43.5\%, 54.3\% and 60.8\% for the new operators, respectively. The LSTM model clearly underperformed compared to the other two approaches. As expected, both the Transformer and xLSTM achieved satisfactory results for the operator they were trained on, though the xLSTM model demonstrated better generalization capabilities to new operators. The results clearly show that for this type of classification, the xLSTM model offers a slight edge over Transformers.
\end{abstract}

% keywords can be removed

\section{Introduction}
\label{sec1}

% (h-closest to fig. ref in text, b-bottom of page, t-top of page)
\begin{figure*}[ht]
	\includegraphics[width=1\textwidth]{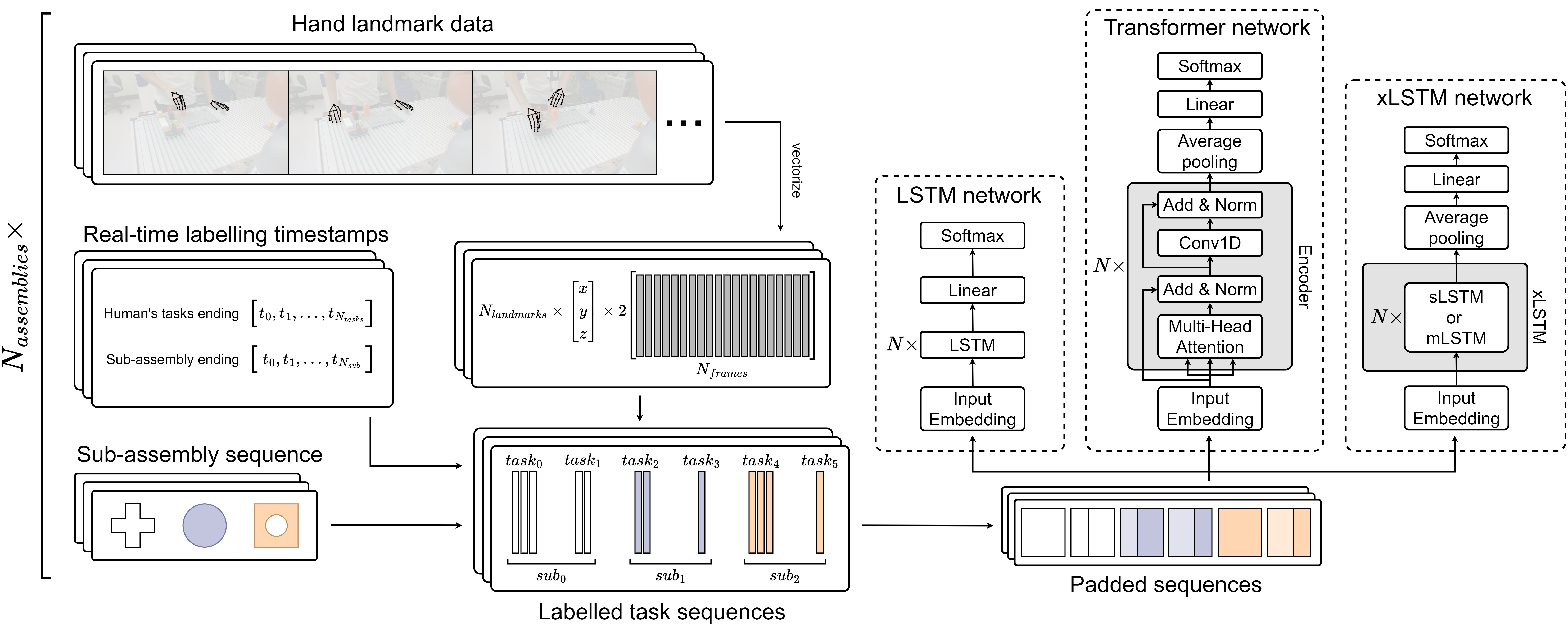}
	\caption{Training Pipeline. Extracted landmark coordinates are divided into sequences and padded according to their unique labels. Finally, the padded sequences are used to train the LSTM, Transformer, and xLSTM models.}
	\label{fig:pipeline}
\end{figure*}

Advanced robotic systems must understand human actions to (1) learn collaboratively, (2) anticipate human needs, (3) ensure safety, and (4) adapt flexibly to increasingly unstructured environments \cite{Wang2019,Mukherjee2022,Bi2021}. In this context, the classification of human-performed assembly tasks is crucial for advancing collaborative robotics across all domains, particularly in manufacturing \cite{Duarte2024,Fan2022,Schou2018}. Most tasks performed by humans consist of sequential primitive actions, varying in length and complexity. While some tasks may seem trivial to classify using today’s methods, the classification of long-horizon tasks remains a significant challenge. The accuracy of human action classification depends on factors such as the length of the action sequence, as well as the number and similarity of the different primitive tasks it includes. Segmenting long tasks into smaller primitive actions is often challenging or unfeasible, highlighting the need for classifying relatively long tasks that encompass multiple primitive actions. For robots to recognize human actions, they require advanced perception and reasoning capabilities, supported by powerful machine learning models \cite{Buerkle2022,Male2022,Grigore2018}. The recognition of human intent and the impact of robot movement on overall coordination are key factors \cite{Chang2018}. An important criterion for the human-robot collaborative process is the signal timing used to deliver collaboration cues, as wrongly timed signals can lead to a deterioration in coordination \cite{Cini2021,Wang2022}. A recent study proposes to solve long sequential tasks over long time horizons in robotic manipulation skills \cite{triantafyllidis2023}.

In the last two decades, a substantial number of classifiers for handling sequential data have relied on long short-term memory (LSTM) networks \cite{Hochreiter1997}, powered by data from different sensors to recognize gestures, motion primitives or primitive manufacturing tasks \cite{Simao2019,Chen2021,Duarte2023}. These recurrent networks have been widely studied and applied, achieving significant performance in classifying time series patterns. However, for longer sequences, these networks exhibit some performance deterioration.

In 2017, Transformer networks revolutionized the field of artificial intelligence, significantly improving the performance of classification for longer pattern sequences \cite{Vaswani2017}. Transformer-based networks have demonstrated highly promising results across diverse fields and have been shown to outperform LSTM-based solutions in tasks such as horse behaviour recognition, action recognition, and intent prediction \cite{Martin-Cirera2024,Girdhar2019}. However, a study on online trading has shown that LSTM networks may still outperform Transformer-based methods in certain tasks \cite{Bilokon2023}. The success of Transformer-based networks in large language models (LLMs) has motivated their application to human action recognition \cite{Ulhaq2022,Xin2023,Girdhar2019}. Studies have also demonstrated promising results in applying Transformer-based networks to both action recognition and action anticipation tasks \cite{Wang2021,Gong2022}. Additionally, other research has focused on recognizing various aspects of human-related information, such as intent \cite{Chang2018,Kedia_2024}, state \cite{Buerkle2022}, and task anticipation \cite{forlini}. 

In 2024, the extended Long Short-Term Memory (xLSTM) architecture emerged as a promising alternative to Transformers for processing temporal data \cite{Beck2024}. By introducing exponential gating and enhanced memory structures, xLSTM addresses limitations of traditional LSTMs, such as limited storage capacity and lack of parallelizability, thereby improving the modelling of long-term dependencies in sequential data. Recent results suggest that xLSTM-based models provide a competitive alternative to Transformer-based models for long-term time series forecasting tasks \cite{Alharthi2024}.

In this study, we propose classifying long assembly sequential tasks that include multiple primitive actions using hand landmark coordinate data from the CT benchmark dataset, Fig.~\ref{fig:pipeline}. The performance of two well-established classifiers, LSTMs and Transformers, is compared with the recent xLSTM model. Additionally, we propose a novel methodology for real-time labelling of different assemblies, reducing the human effort and time required to set up and operate the system.

\global\long\def\S{\mathcal{S}}
\global\long\def\R{\mathbb{R}}

%--------------------------------------------------------------------------------------------
\section{Methodology} %Materials and Setup
\label{sec:3}
\subsection{Long Short-Term Memory (LSTM)}
\label{sec:3.1}

Recurrent neural networks (RNNs) were developed from standard feed forward neural networks (FFNN) by adding recurrent connections and a hidden state vector, enabling them to capture time-relationships within sequences of inputs. Though, vanilla RNNs often struggle in retaining long-term dependencies \cite{Bengio1994}. This limitation led to the development of LSTM networks \cite{Hochreiter1997}. LSTMs have two key components: the cell state \(c\) and the hidden state \(h\). The cell state is the memory of the network, carrying information across different time steps, i.e., long-term memory, and the hidden state is the output of the LSTM cell for a given time step, i.e., short-term memory. These states are updated and managed using three gates: the forget gate, the input gate, and the output gate.

The forget gate determines what information to discard from the previous cell state through the following:
\begin{equation}
\label{eq:1}
f_t = \sigma \left(\mathbf{w}_f^{\mathsf{T}} \mathbf{x}_t + r_f h_{t-1} + b_f\right)
\end{equation}
Where \(f_t\) is the forget gate at time step \(t\), \(\sigma\) is the Sigmoid activation function, \(\mathbf{w}_f\) is the weight vector between the input vector \(\mathbf{x}_t\) and the forget gate, \(h_{t-1}\) is the hidden state from the previous time step, \(r_f\) are the recurrent weights between the hidden state and the forget gate, and \(b_f\) is the bias vector for the forget gate.

The input gate determines what information to add to the cell state. First a candidate state is calculated from:
\begin{equation}
\label{eq:2}
\tilde{c}_t = \textrm{tanh}\left(\mathbf{w}_c^\mathsf{T} \mathbf{x}_t + r_c h_{t-1} + b_c\right) 
\end{equation}
Where \(\tilde{c}_t\) is the candidate cell state, tanh is the hyperbolic tangent activation function, \(\mathbf{w}_c\) is the weight vector between the input \(\mathbf{x}_t\) and the candidate cell state, \(r_c\) are the recurrent weights between the hidden state and the candidate cell state and \(b_c\) is the bias vector for the candidate cell state. Then, the amount by which the cell state is updated with the candidate cell state is calculated as follows:
\begin{equation}
\label{eq:3}
i_t = \sigma \left(\mathbf{w}_i^\mathsf{T} \mathbf{x}_t + r_i h_{t-1} + b_i\right)
\end{equation} 
Where \(i_t\) is the input gate activation vector, \(\mathbf{w}_i\) is the weight vector between the input \(\mathbf{x}_t\) and the input gate, \(r_i\) are the recurrent weights between the hidden state and the input gate and \(b_i\) is the bias vector for the input gate. Thus, the new cell state is given by:
\begin{equation}
\label{eq:4}
c_t = f_t \cdot c_{t-1} + i_t \cdot {\tilde{c_t}}
\end{equation}

Lastly, the output gate decides what information to output from the current cell state through:
\begin{equation}
\label{eq:5}
o_t = \sigma \left(\mathbf{w}_o^\mathsf{T} \mathbf{x}_t + r_o h_{t-1} + b_o\right)
\end{equation}
Where \(o_t\) is the output gate activation vector, \(\mathbf{w}_o\) is the weight vector between the input \(\mathbf{x}_t\) and the output gate, \(r_o\) are the recurrent weights between the hidden state and the output gate and \(b_o\) is the bias vector for the output gate. The hidden state is then calculated by:
\begin{equation}
\label{eq:6}
h_t = o_t \cdot \textrm{tanh}\left(c_t\right)
\end{equation}

\subsection{Transformers}
\label{sec:3.2}

The Transformer architecture \cite{Vaswani2017} is built upon the attention mechanism introduced by \cite{Bahdanau2016}. The main idea of the attention mechanism is to enable the model to focus on different parts of the input sequence when producing the output. Its key components are the query \(\mathbf{Q}\), key \(\mathbf{K}\) and value \(\mathbf{V}\), which represent, respectively, the current item or state being focused on, the reference items we compare against the query to determine relevance, and the values associated with each key. The attention scores are then calculated through:
\begin{equation}
\label{eq:7}
\textrm{Attention scores} = \textrm{softmax}\left(\frac{\mathbf{QK}^T}{\sqrt{d_k}}\right)\mathbf{V}
\end{equation}
Where \(d_k\) is the dimension of keys and queries.

The Transformer architecture leverages the usage of Multi-Head Attention, where instead of computing a single set of attention scores, the queries, keys, and values are split into multiple heads and attention scores are computed for each head. The results are then concatenated and linearly transformed with \(\mathbf{W}_i^Q\), \(\mathbf{W}_i^K\), \(\mathbf{W}_i^V\) being the projection matrices that transform the input into the query, key and value vectors for head \(i\), and \(\mathbf{W}^O\) the output projection matrix:
\begin{equation}
\label{eq:8}
\textrm{MultiHead}\left(\mathbf{Q}, \mathbf{K}, \mathbf{V}\right) = \textrm{Concat}\left(head_1, ..., head_n\right)\mathbf{W}^O
\end{equation}
Where:
\begin{equation}
\label{eq:8.5}
\textrm{head}_i = \textrm{Attention}\left(\mathbf{QW}_i^Q, \mathbf{KW}_i^K, \mathbf{VW}_i^V\right)
\end{equation}

The Transformer architecture proposed in \cite{Vaswani2017} has an encoder and a decoder. The encoder is comprised of \(N\) stacks of two sub-layers: a multi-head self-attention mechanism and a position-wise FFNN. The decoder architecture also has \(N\) stacks of three sub-layers: a masked multi-head self-attention mechanism, a multi-head self-attention mechanism and a position-wise FFNN. In both the encoder and decoder, each sub-layer has a residual connection followed by layer normalisation. Though, for the purpose of pattern classification, the transformer network only requires the encoder, followed by some fully connected layers and a softmax layer.

% (h-closest to fig. ref in text, b-bottom of page, t-top of page)
\begin{figure*}[ht]
	\includegraphics[width=1\textwidth]{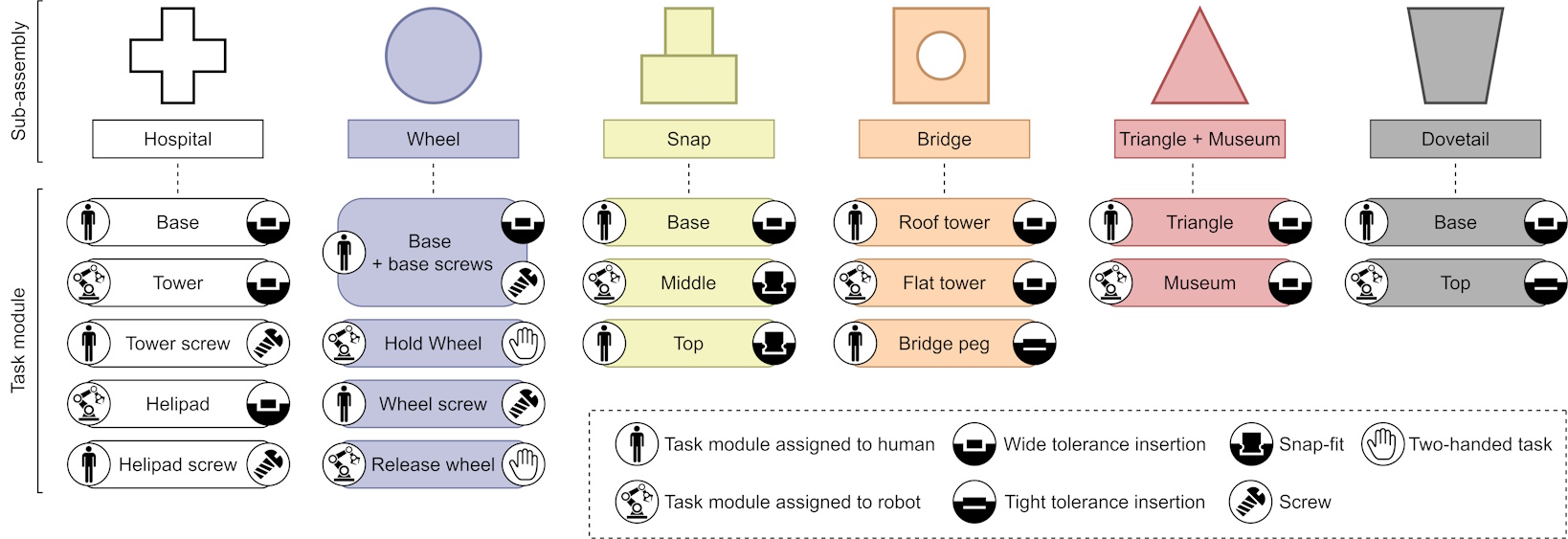}
	\caption{CT benchmark sub-assemblies and their respective sequence of task modules. Each task module is assigned to an agent (human or robot), represented by the symbol on the left, and contains one or more tasks, indicated by the symbol on the right.}
	\label{fig:sequence}
\end{figure*}

\subsection{xLSTM}
\label{sec:3.3}
LSTM networks suffer from some limitations, mainly the inability to revise storage decisions, the limited storage capabilities and the lack of parallelisation. For these reasons, the xLSTM \cite{Beck2024} was proposed as an extension to the LSTM architecture.
This new architecture introduces two main modifications: exponential gating and novel memory structures. From these modifications, two new blocks were proposed, the sLSTM and the mLSTM. An xLSTM network is comprised of both sLSTM and mLSTM blocks.

\subsubsection{sLSTM}
\label{sec:3.3.1}
In the sLSTM block, exponential gating is applied to the input gate as follows:
\begin{equation}
\label{eq:9}
i_t = \exp{\left(\mathbf{w}_i^\mathsf{T} \mathbf{x}_t + r_i h_{t-1} + b_i\right)}
\end{equation}
Where \(\exp\) is the exponential activation function. Optionally, exponential gating can also be applied to the forget gate as follows: 
\begin{equation}
\label{eq:10}
f_t = \sigma{\left(\mathbf{w}_f^\mathsf{T} \mathbf{x}_t + r_f h_{t-1} + b_f\right)} \textrm{ or }\exp{\left(\mathbf{w}_f^\mathsf{T} \mathbf{x}_t + r_f h_{t-1} + b_f\right)}
\end{equation}
This block has two new states, the normaliser state \(n_t\) and the stabiliser state \(m_t\). The later was introduced to prevent overflows due to exponential gating. The normaliser state is calculated by:
\begin{equation}
\label{eq:11}
n_t = f_t \cdot n_{t-1} + i_t
\end{equation}
It is used to calculate the hidden state as follows:
\begin{equation}
\label{eq:12}
h_t = o_t \cdot \textrm{tanh}\left(\frac{c_t}{n_t}\right)
\end{equation}
The stabiliser state is calculated through:
\begin{equation}
\label{eq:13}
m_t=\max\left(\log\left(f_t\right)+m_{t-1},\log\left(i_t\right)\right)
\end{equation}
This is used to stabilise the input gate:
\begin{equation}
\label{eq:14}
i'_t=\exp\left(\log\left(i_t\right)-m_t\right)=\exp\left(\mathbf{w}_i^\mathsf{T} \mathbf{x}_t + r_i h_{t-1} + b_i - m_t\right)
\end{equation}
And the forget gate:
\begin{equation}
\label{eq:15}
f'_t=\exp\left(\log\left(f_t\right)+m_{t-1}-m_t\right)
\end{equation}

Apart from these modifications, the sLSTM block can have multiple memory cells, similar to the original LSTM architecture. Having multiple cells enables memory mixing through recurrent connections. As such, sLSTM blocks can have multiple heads with memory mixing within each head, but not across heads.

\subsubsection{mLSTM}
\label{sec:3.3.2}
To improve the storage capabilities of LSTM networks, the memory cell in the mLSTM block was increased from a scalar \(c\in \mathbb{R}\) to a matrix \(\mathbf{C}\in \mathbb{R}^{d\times d}\). This block has three input vectors, the query \(\mathbf{q}_t\), the key \(\mathbf{k}_t\), and the value \(\mathbf{v}_t\), which are defined respectively as follows:
\begin{equation}
\label{eq:16}
\mathbf{q}_t=\mathbf{W}_q\mathbf{x}_t+\mathbf{b}_q
\end{equation}
Where \(\mathbf{W}_q\) is the weight matrix for the query and \(\mathbf{b}_q\) is the bias vector for the query: 
\begin{equation}
\label{eq:17}
\mathbf{k}_t=\frac{1}{\sqrt{d}}\mathbf{W}_k\mathbf{x}_t+\mathbf{b}_k
\end{equation}
Where \(d\) is the vector dimension, \(\mathbf{W}_k\) is the weight matrix for the key, and \(\mathbf{b}_k\) is the bias vector for the key:
\begin{equation}
\label{eq:18}
\mathbf{v}_t=\mathbf{W}_v\mathbf{x}_t+\mathbf{b}_v
\end{equation}
Where \(\mathbf{W}_v\) is the weight matrix for the value and \(\mathbf{b}_v\) is the bias vector for the value.

Due to the introduction of three new input vectors, all other equations were redefined. The new input gate, forget gate, output gate, cell state, normaliser state, hidden state, are as follows:
\begin{equation}
\label{eq:19}
i_t = \exp (\mathbf{w}_i^\mathsf{T} \mathbf{x}_t + b_i)
\end{equation}
\begin{equation}
\label{eq:20}
f_t = \sigma (\mathbf{w}_f^\mathsf{T} \mathbf{x}_t + b_f) \textrm{ OR }\exp (\mathbf{w}_f^\mathsf{T} \mathbf{x}_t + b_f)
\end{equation}
\begin{equation}
\label{eq:21}
\mathbf{o}_t = \sigma (\mathbf{W}_o \mathbf{x}_t + \mathbf{b}_o)
\end{equation}
Where \(\mathbf{W}_o\) is the weight matrix between the input \(\mathbf{x}_t\) and the output gate:
\begin{equation}
\label{eq:22}
\mathbf{C}_t=f_t\cdot \mathbf{C}_{t-1} + i_t\cdot \mathbf{v}_t\mathbf{k}_t^\mathsf{T}
\end{equation}
\begin{equation}
\label{eq:23}
\mathbf{n}_t = f_t \cdot \mathbf{n}_{t-1} + i_t\cdot \mathbf{k}_t
\end{equation}
\begin{equation}
\label{eq:24}
\mathbf{h}_t = \mathbf{o}_t \odot \left(\frac{\mathbf{C}_t\mathbf{q}_t}{\max\left\{\mid \mathbf{n}_t^\mathsf{T} \mathbf{q}_t\mid,1\right\}}\right)
\end{equation}

\subsection{Collaborative assembly scenario}
\label{sec:3.4}
The human-robot collaborative assembly scenario, based on the CT benchmark, involves assembling a city landscape model \cite{Duarte2024}. For a more rigorous experiment, the assembly hierarchy is defined as follows (from lower to higher levels): primitive task, task, task module, sub-assembly, and assembly.
Primitive tasks are short, distinguishable actions, such as picking, placing, or holding. However, segmenting these lower-level actions is often impractical. A combination of primitive tasks forms a task, such as insertion or screw fastening.
A task module comprises one or more tasks associated with a specific component and assigned to an agent (human or robot). A group of task modules constitutes a sub-assembly, and all sub-assemblies together form the complete assembly. 

The studied HRC scenario is comprised of six sub-assemblies, each one related to a different building, Fig.~\ref{fig:sequence}. These sub-assemblies are the long assembly sequential tasks we aim to classify, by observing data corresponding to one of its task modules. Each sub-assembly consists of two to five task modules, with one to three assigned to the human.

% (h-closest to fig. ref in text, b-bottom of page, t-top of page)
\begin{figure*}[!ht]
	\centering\includegraphics[width=0.9\textwidth]{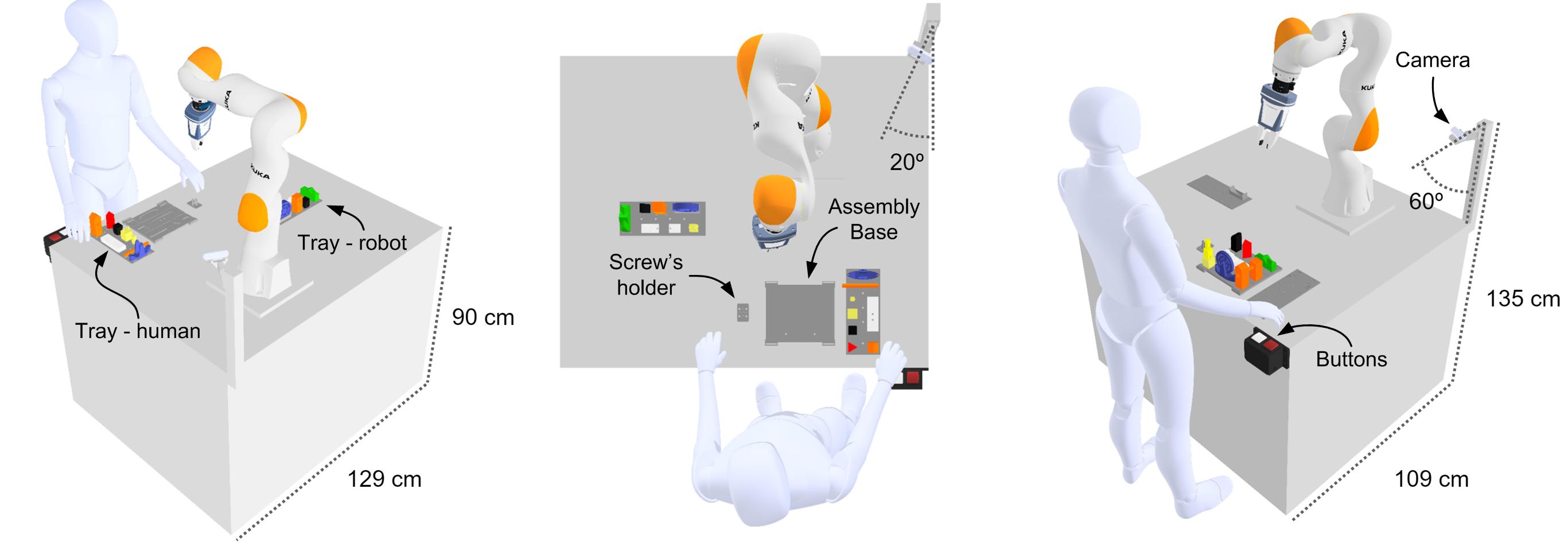}
	\caption{Multiple views of the assembly layout showcasing the arrangement of elements and their relative positions.}
	\label{fig:layout}
\end{figure*}

The landscape models are arranged in two trays, with the human hands positioned within the camera field of view, Fig.~\ref{fig:layout}. The tray holding the parts to be assembled by the human, along with the screw holder, is placed next to the assembly base. The tray containing the parts for the robot assembly is positioned farther from the human but remains within the robot reach.

\begin{figure}[!ht]
	\centering\includegraphics[width=1\textwidth]{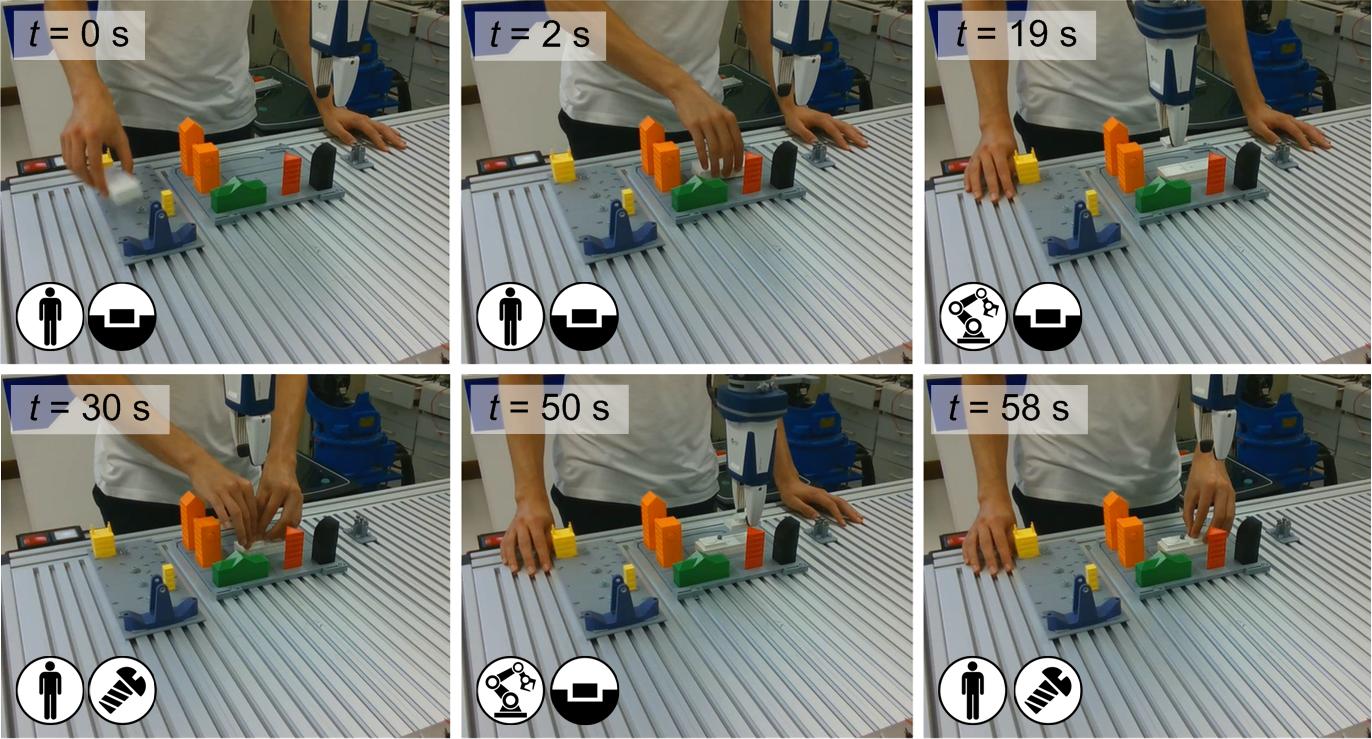}
	\caption{Sub-assembly of the hospital building, comprising task modules performed by both the human and the robot. The entire process took approximately 60 seconds to complete.}
	\label{fig:hospital_assembly}
\end{figure}

During the assembly process, RGB footage was captured using an Intel RealSense D435i at a frame rate of 10 fps and a resolution of 1280 × 720 pixels. Real-time labelling was conducted by the human using the two available buttons, significantly reducing post-labelling effort. Prior to each assembly, the human operator provided the sequence of sub-assemblies in advance. During the process, the operator used the buttons to indicate the start of task models performed by the robot and to mark the completion of each sub-assembly. Fig.~\ref{fig:hospital_assembly} illustrates the sub-assembly of the hospital building, comprising task modules performed by both the human and the robot. The entire process took approximately 60 seconds to complete.

%---------------------

% (h-closest to fig. ref in text, b-bottom of page, t-top of page)

\subsection{Data padding and augmentation}
\label{sec:3.5}

While acquiring RGB data from the assemblies, the hand positions (landmarks) were extracted through mediapipe \cite{Lugaresi2019}. Then, the gathered hand positions were compiled in a single vector of \(x\), \(y\), \(z\) coordinates for each 21 landmarks and for each hand, accounting for a total vector size of 126. According to the gathered sub-assembly labels, landmark vectors were separated into task modules, which were padded to have the same length as the longest recorded task module. In total, four padding techniques were employed: zero, idle, random and real. For zero-padding, the sequences were padded with arrays of zeros. For idle padding, the sequences were padded randomly with data corresponding to an idle state, i.e. waiting for the robot. For the random padding, sequences were padded with the end of any other sequence randomly. For the real padding, the sequences were padded with the data that actually preceded it. These task module sequences were then used to train a classification model to map the observed task model to the sub-assembly it belonged to, Fig.~\ref{fig:pipeline}.

To improve the generalisation capabilities of the models, during training two augmentation techniques were applied: temporal resize and noise. All of these techniques were applied partially and randomly to the dataset along the epochs. For the temporal resize, the sequences were either shortened or lengthened through linear interpolation. For the noise augmentation, random noise was applied along the sequence to produce a slightly different trajectory. Both techniques were implemented with normal distributions.

\section{Results and discussion}
\label{sec:4}

\begin{figure}[!ht]
	\centering\includegraphics[width=1\textwidth]{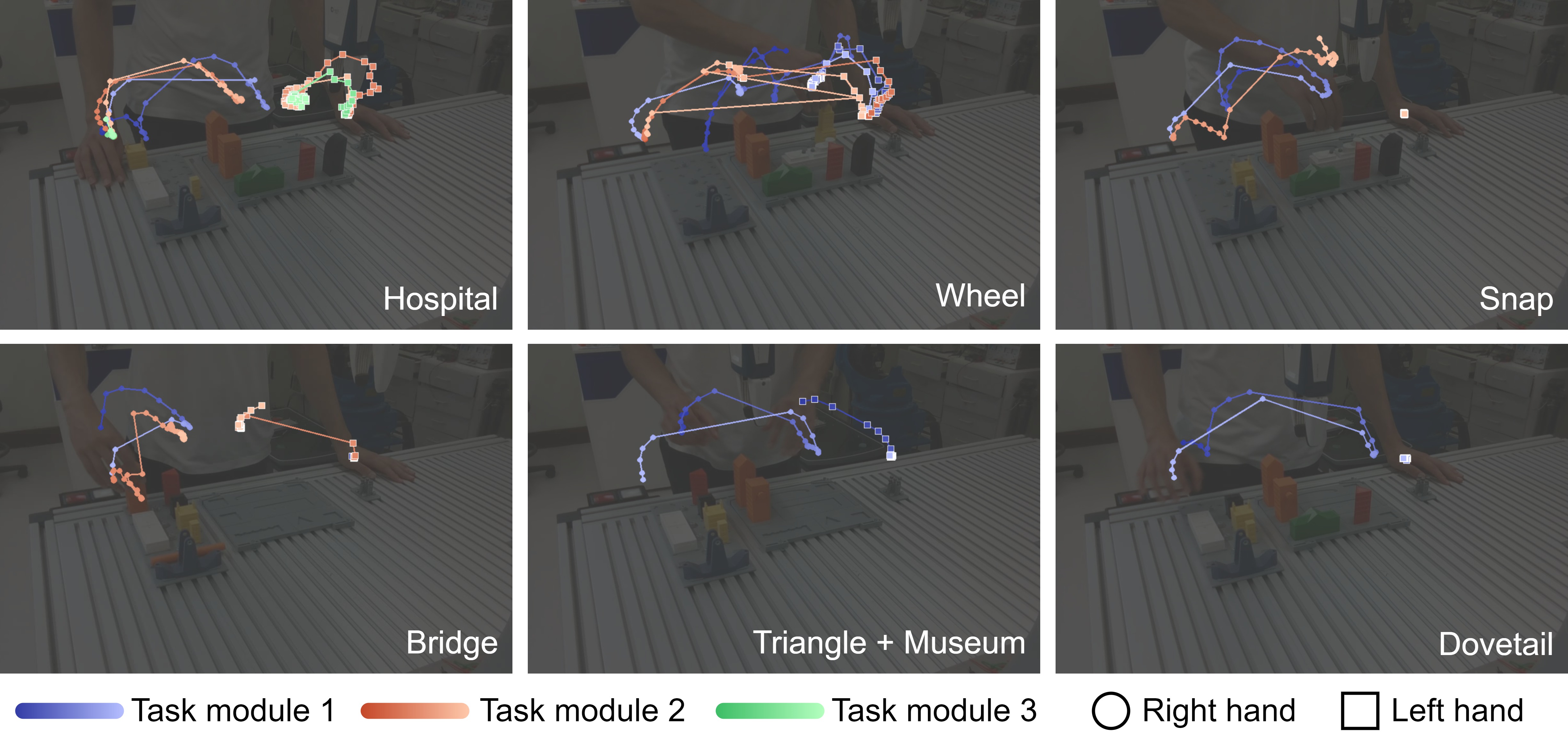}
	\caption{Trajectory samples for each sub-assembly. The direction of the hand’s trajectory is represented by a gradient from darker to lighter colours. Right-hand trajectories are depicted with circles, while left-hand trajectories are shown as squares with white outlines. The colours indicate the order of human task modules in the sub-assembly.}
	\label{fig:trajectory}
\end{figure}

A total of 70 full assemblies were performed and recorded by four operators. Of these, 55 assemblies were completed by a single operator (operator 1), while the remaining 15 assemblies were evenly distributed among the other three operators (operators 2, 3, and 4). Trajectory samples for each sub-assembly are shown in Fig.~\ref{fig:trajectory}.

The data was divided into training (40 assemblies), validation (10 assemblies), and testing sets (20 assemblies). The training and validation sets consisted solely of assemblies performed by operator 1, while the testing set included 5 assemblies from each operator. The labeled sequences of each assembly were padded to match the longest sequence, which was 186 frames (18.6 seconds). 

All architectures were trained in a computer with a Nvidia GeForce RTX 3070 Ti GPU (8GB VRAM), an AMD Ryzen 9 3900X CPU and 32GB of RAM. The LSTM and Transformer architectures were implemented with Tensorflow, and the xLSTM architecture was implemented with Pytorch.
Due to the limited number of assemblies, results for each experiment were obtained using a five-fold cross-validation process, with the hyperparameters of all architectures carefully optimized. In an initial study, the models were trained, validated, and tested using zero-padded sequences. A hyperparameter search was conducted, and the best-performing values are summarized in Table~\ref{tab:hyperparameters}. The mean training times of each cross-validation model for the LSTM, Transformer and xLSTM architectures were 15m 50.6s, 5m 12.0s, and 46m 55.4s, respectively. Additionally, the average inference time was 1.6ms for the LSTM, 3.2ms for the Transformer and 4.9ms for the xLSTM.

\begin{table*}[!ht]
	\centering
	\caption{Best performing hyperparameters for training and validation with zero-padded sequences.}
	\label{tab:hyperparameters}       % Give a unique label
	\begin{tabular}{llllll}
		\hline\noalign{\smallskip}
		\multicolumn{2}{c}{LSTM}  & \multicolumn{2}{c}{Transformer}             & \multicolumn{2}{c}{xLSTM}\\
		\noalign{\smallskip}\hline\noalign{\smallskip}
		\multicolumn{6}{c}{Training parameters} \\
		\noalign{\smallskip}\hline\noalign{\smallskip}
        epochs             & 500  & epochs              & 1000   & epochs             & 300      \\
		batch size         & 64   & batch size          & 32     & batch size         & 32       \\
		learning rate      & 1e-5 & learning rate       & 2.5e-5 & learning rate      & 2.5e-5   \\
		noise probability  & 0    & noise probability   & 0.3    & noise probability  & 0.2      \\
		resize probability & 0    & resize probability  & 0.3    & resize probability & 0        \\
                           &      & noise std           & 0.05   & noise std          & 0.05     \\
                           &      & resize std          & 0.1    &                    &          \\
		\noalign{\smallskip}\hline\noalign{\smallskip}
		\multicolumn{6}{c}{Architecture specifics} \\
		\noalign{\smallskip}\hline\noalign{\smallskip}
		dropout            & 0    & dropout             & 0.2    & dropout            & 0        \\
		recurrent dropout  & 0    & number of heads     & 4      & number of heads    & 4        \\
		units              & 256  & head dimension      & 256    & head dimension     & 256      \\
                           &      & number of blocks    & 2      & number of blocks   & 7        \\
                           &      & Conv 1D filter size & 2      & sLSTM locations    & 3rd, 5th \\
		\noalign{\smallskip}\hline\noalign{\smallskip}
		\multicolumn{6}{c}{Output parameters} \\
		\noalign{\smallskip}\hline\noalign{\smallskip}
		dense units        & 128  & dense units         & 128    & dense units        & 128      \\
		dense dropout      & 0    & dense dropout       & 0      & dense dropout      & 0        \\
		\noalign{\smallskip}\hline
	\end{tabular}
\end{table*}

\begin{figure}[!ht]
	\centering\includegraphics[width=1\textwidth]{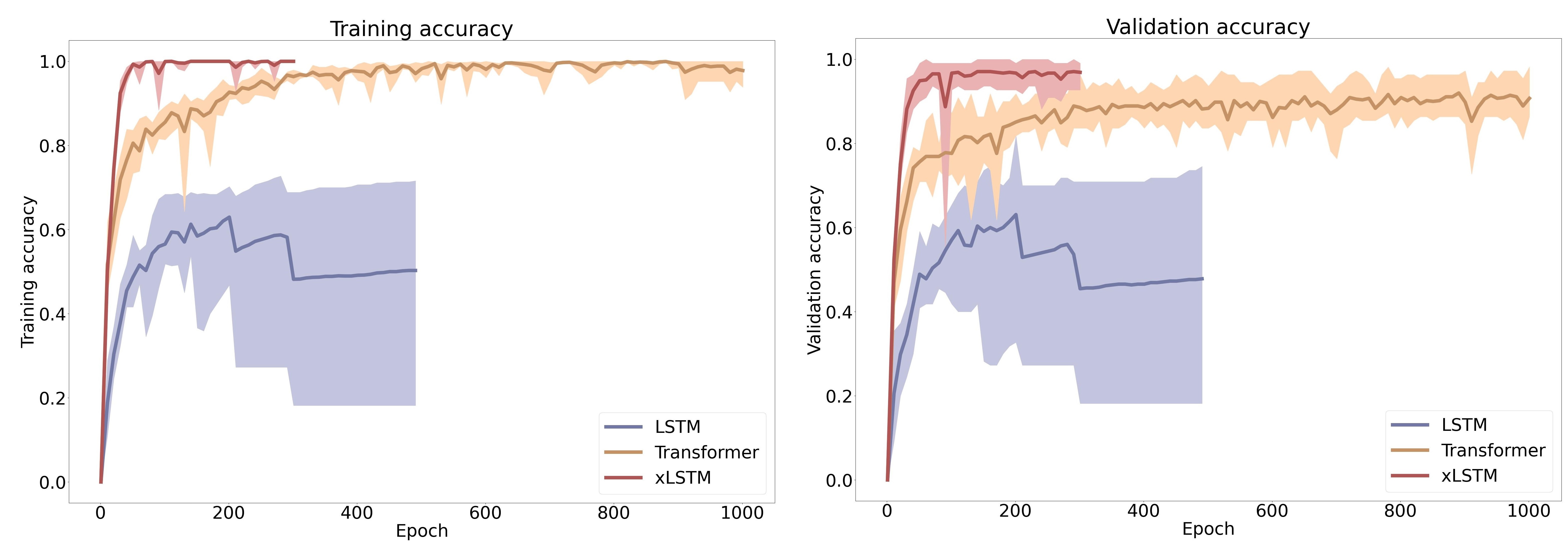}
	\caption{Training and validation accuracies for best performing models of the LSTM, Transformer and xLSTM architectures.}
	\label{fig:6}
\end{figure}

The evolution of the average, minimum, and maximum training and validation accuracies across the five-fold cross-validation models is shown in Fig.~\ref{fig:6}. In an initial analysis, both the Transformer and xLSTM architectures achieved higher accuracies compared to the LSTM architecture and exhibited more stable learning curves. During the hyperparameter search, the LSTM models struggled to learn consistently, underscoring their limitations in classifying long temporal sequences. Furthermore, the xLSTM model required fewer epochs to reach peak performance and achieved overall better validation accuracies than the Transformer model. For all architectures, the model weights used for testing corresponded to those achieving the best results during the training and validation stages.

% (h-closest to fig. ref in text, b-bottom of page, t-top of page)

For the testing results, two main accuracies were evaluated: the testing accuracy for all five assemblies for each operator and the testing accuracy for operators who did not provide assemblies for the testing and validation sets. As expected, the testing assemblies performed by the operator who provided the train and validations sets (operator 1) were more accurately classified. The testing accuracies for all operators and for the purely testing operators were \(59.1\%\) and \(54.1\%\) for the LSTM model, \(75.1\%\) and \(67.2\%\) for the Transformer model, and \(80.3\%\) and \(74.6\%\) for the xLSTM model, respectively. A more detailed analysis of the testing accuracies can be observed in Table~\ref{tab:accuracies}. Both the Transformer and xLSTM models outperform the LSTM model. Additionally, the xLSTM outperforms the Transformer model in accuracy for operators 3 and 4. Another observation is that assemblies recorded by operator 4 achieved significantly better results compared to those performed by operators 2 and 3, even though none of these operators contributed to the training and validation sets. This difference may be attributed to hand size, as the approximate hand lengths for operators 1, 2, 3, and 4 were \(20.5\,\)cm, \(17.0\,\)cm, \(18.0\,\)cm and \(20.0\,\)cm, respectively. As such, to improve the robustness of the system across different operators, the training and validation sets should contain assemblies from operators with diverse hand sizes.

\begin{table}[!ht]
	\centering
	\caption{Average testing accuracy for each assembly of each operator. The highest average accuracy per operator is highlighted in bold, while the second-best accuracy is underlined.}
	\label{tab:accuracies}       % Give a unique label
	\begin{tabular}{ccccccc}
		\noalign{\smallskip}\hline
                    & Op. 1            & Op. 2            & Op. 3            & Op. 4            & Average \\
		\noalign{\smallskip}\hline\noalign{\smallskip}
		LSTM        & 74.0             & 47.4             & 41.7             & 73.2             & 59.1    \\
		Transformer & \textbf{98.7}    & \underline{61.8} & \underline{52.2} & \underline{87.6} & \underline{75.1}    \\
		xLSTM       & \underline{97.3} & \textbf{63.6}    & \textbf{63.7}    & \textbf{96.4}    & \textbf{80.3}    \\
		\noalign{\smallskip}\hline
	\end{tabular}
\end{table}

\begin{figure*}[!ht]
	\centering\includegraphics[width=1\textwidth]{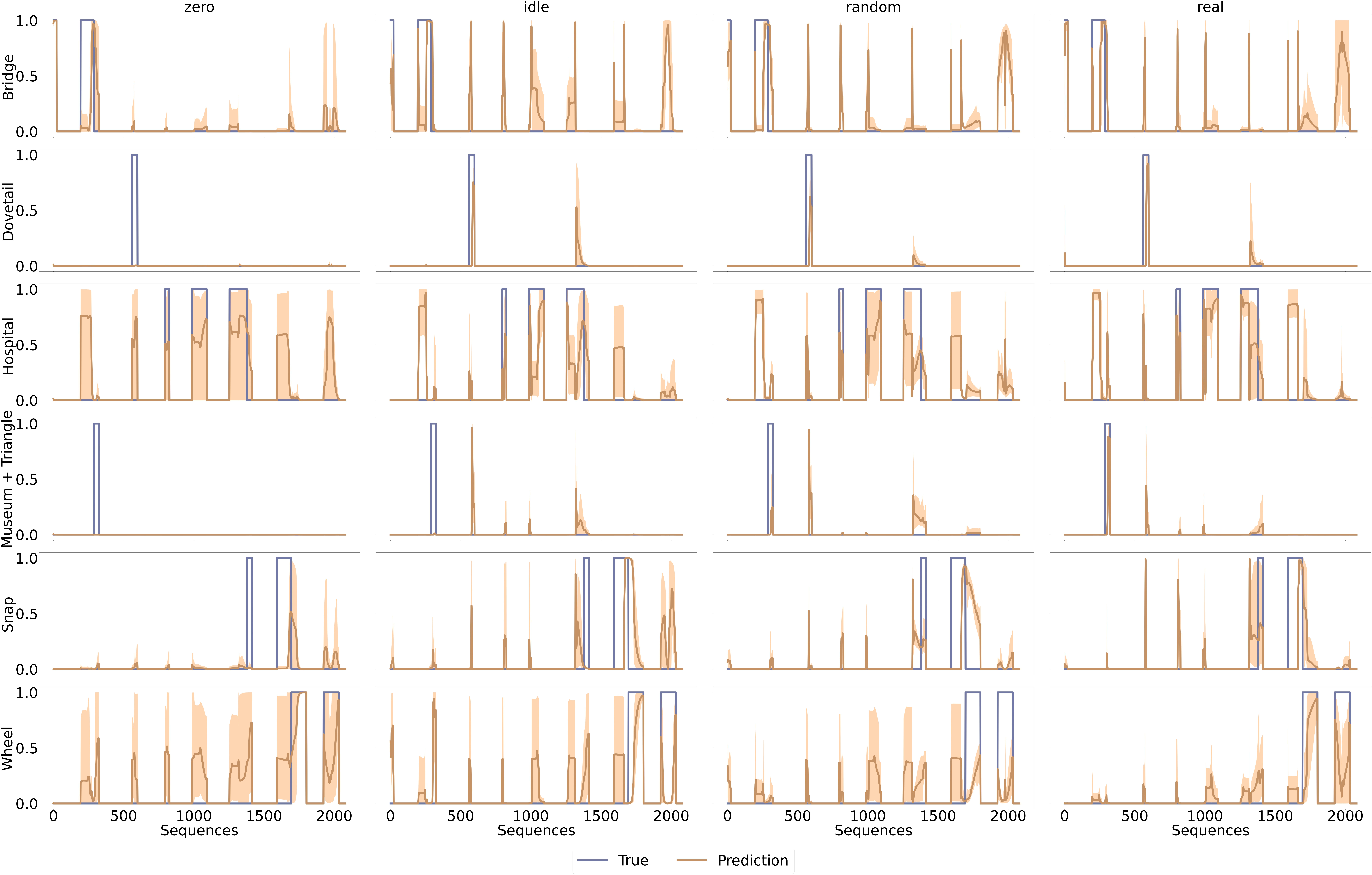}
	\caption{True labels and predictions from cross-validation models for a testing video performed by operator 1 are presented for all four padding techniques (zero-padding, idle-padding, random-padding, and real-padding) using the Transformer model as the classifier.}
	\label{fig:padding_real_time}
\end{figure*}

Although the models achieved satisfactory results when tested with zero-padding, this performance dropped during real-time testing, with a continuous data stream as input. For an illustrative comparison, we evaluated these models using real-padded sequences, which closely approximate real-time conditions, and the resulting accuracies for the LSTM, Transformer, and xLSTM models were \(37.5\%\), \(28.6\%\) and \(31.5\%\), respectively. These results suggest the zero-padding approach for training is a suboptimal technique as it may not effectively generalise to real-time scenarios. Thus, the four padding methods were compared using the Transformer model with the previously selected hyperparameters, Table~\ref{tab:hyperparameters}. The resulting testing accuracies for models trained with zero-padded, idle-padded, random-padded and real-padded sequences were \(34.3\%\), \(54.3\%\), \(60.3\%\) and \(62.1\%\), respectively. The four padding techniques were also compared for real-time classification, using a sliding window approach with a stride of 1, for one of the testing assemblies performed by operator 1, Fig.~\ref{fig:padding_real_time}. For sequences where the robot was performing a task module the predictions were not deemed relevant, and, as such, in the plots the model predictions were displayed as 0 for all labels.

 % (h-closest to fig. ref in text, b-bottom of page, t-top of page)

\begin{figure}[!ht]
	\centering\includegraphics[width=1\textwidth]{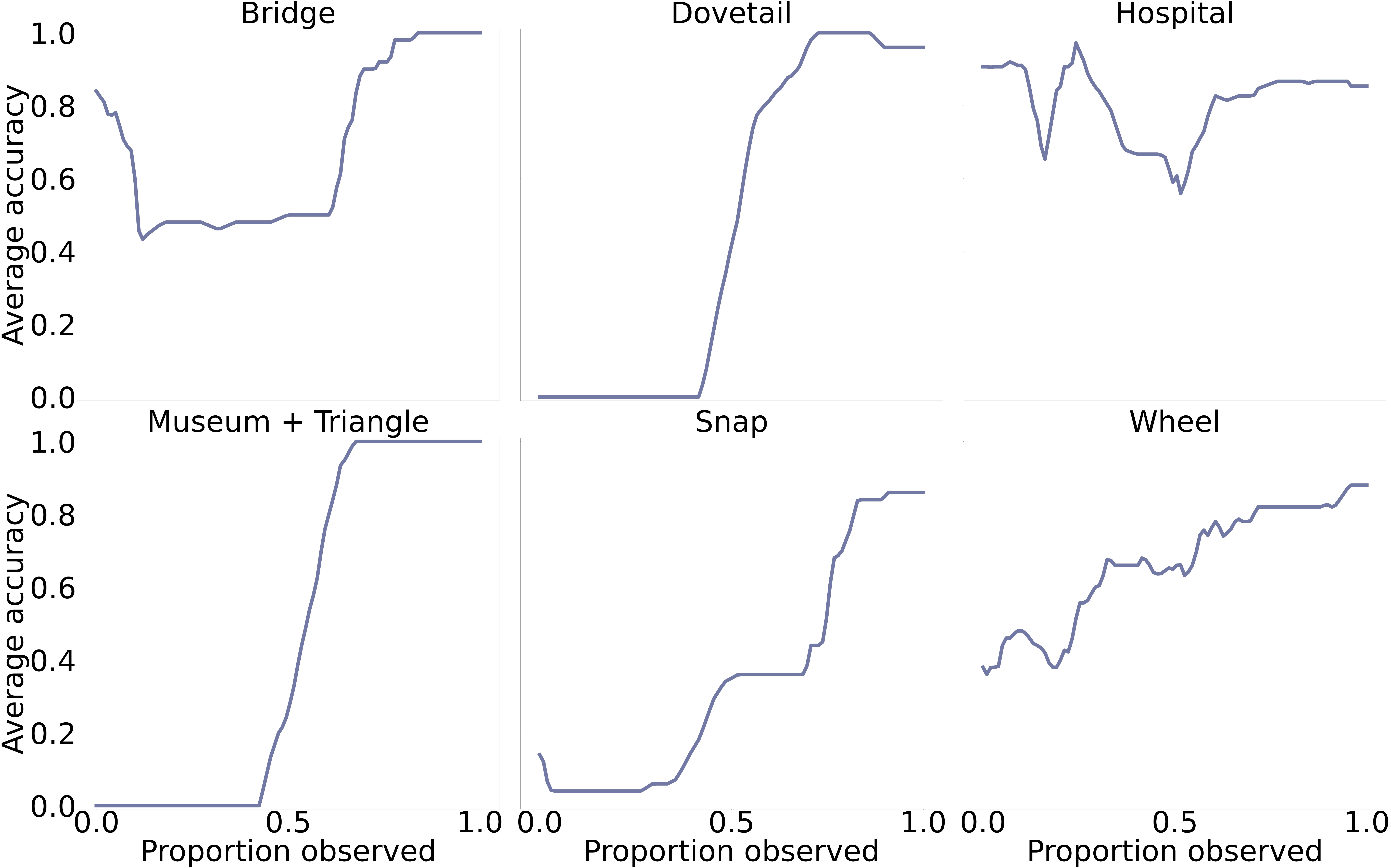}
	\caption{Accuracy evolution relative to the proportion of the task module observed for operator 1, using the Transformer network trained with real-padded sequences.}
	\label{fig:accuracy_evolution}
\end{figure}

The average testing results suggest that real-padding is the best-performing approach for transitioning to real-time classification. This is further supported by the real-time classification plots, which show that shorter sequences are more frequently classified correctly, and that performance is more robust across the cross-validation models. The incorrect spikes observed during real-time classification occur because model accuracy improves as more frames from the current task module are observed. This trend is further confirmed by analysing the evolution of accuracy relative to the proportion of the task module observed, Fig.~\ref{fig:accuracy_evolution}.

 % (h-closest to fig. ref in text, b-bottom of page, t-top of page)

Each architecture was trained and validated using real-padded sequences, with a new hyperparameter search conducted, Table~\ref{tab:real_hyperparameters}. The testing accuracies for each architecture are summarized in Table~\ref{tab:real_accuracies}. The average accuracies for all operators and for operators who did not contribute to the training and validation sets were \(50.8\%\) and \(43.5\%\) for the LSTM network, \(64.5\%\) and \(54.3\%\) for the Transformer network and \(68.9\%\) and \(60.8\%\) for the xLSTM network, respectively. Once again, both the Transformer and the xLSTM architectures perform significantly better than the LSTM architecture. The xLSTM architecture presents better generalisation capabilities to new operators (a \(6.5\) percentage points advantage), while the Transformer was slightly more accurate for operator 1 (\(1.8\) percentage points).

\begin{table*}[!ht]
	\centering
	\caption{Best-performing hyperparameters for training and validation using real-padded sequences. Only the hyperparameters that differ from those in Table~\ref{tab:hyperparameters} are displayed.}
	\label{tab:real_hyperparameters}       % Give a unique label
	\begin{tabular}{llllll}
		\hline\noalign{\smallskip}
		\multicolumn{2}{c}{LSTM}   & \multicolumn{2}{c}{Transformer} & \multicolumn{2}{c}{xLSTM}\\
		\noalign{\smallskip}\hline\noalign{\smallskip}
		\multicolumn{6}{c}{Training parameters} \\
		\noalign{\smallskip}\hline\noalign{\smallskip}
		noise probability  & 0.2   & noise probability  & 0.2        & noise probability  & 0.2   \\
		resize probability & 0.2   & resize probability & 0.2        & resize probability & 0.2    \\
		noise std          & 0.025 & noise std          & 0.05       & noise std          & 0.01   \\
		resize std         & 0.025 & resize std         & 0.05       & resize std         & 0.005  \\
		\noalign{\smallskip}\hline\noalign{\smallskip}
		\multicolumn{6}{c}{Architecture specifics} \\
		\noalign{\smallskip}\hline\noalign{\smallskip}
		recurrent dropout  & 0.2   &                    &            & dropout            & 0.2      \\
		\noalign{\smallskip}\hline
	\end{tabular}
\end{table*}

\begin{table}[!ht]
	\centering
	\caption{Average testing accuracy for each assembly of each operator. The highest average accuracy for each operator is highlighted in bold, while the second-highest is underlined.}
	\label{tab:real_accuracies}       % Give a unique label
	\begin{tabular}{cccccc}
		\noalign{\smallskip}\hline
		          & Op. 1            & Op. 2            & Op. 3            & Op. 4            & Average          \\
		\noalign{\smallskip}\hline\noalign{\smallskip}
		LSTM        & 72.9             & 38.4             & 31.4             & 60.6             & 50.8             \\
		Transformer & \textbf{95.0}    & \underline{46.0} & \underline{41.7} & \underline{75.2} & \underline{64.5} \\
		xLSTM       & \underline{93.2} & \textbf{51.4}    & \textbf{51.9}    & \textbf{79.1}    & \textbf{68.9}    \\                       
		\noalign{\smallskip}\hline
	\end{tabular}
\end{table}

The system was tested in real-time classification by Operator 1. The operator had access to a single button, which triggered the model to classify the action being performed and subsequently command the robot to execute the corresponding task module. The attached video demonstrates a successful human-robot assembly process.

\section{Conclusion}

In this study, we analysed and compared the performance of LSTMs, Transformers, and the recent xLSTM model in classifying long-horizon assembly tasks based on hand landmark coordinates. The results, derived from a dataset of assemblies in a HRC setting, highlighted key insights for model selection, training techniques and data preparation in real-time classification. A critical observation for the real-time classification accuracy is the padding technique. For the studied scenario, the real-padding method achieved the best performance when transitioning to real-time classification. Transformer-based and xLSTM-based models clearly outperformed LSTM-based models as the latter tended to be unstable during training and achieved suboptimal classification results. Notably, while Transformers achieved slightly higher accuracy on tasks involving operators included in the training set (\(+1.8\) percentage points), xLSTMs demonstrated superior generalization to new operators (\(+6.5\) percentage points). Future research should explore the performance of these models on larger and more diverse datasets, potentially incorporating additional input signals. Such investigations could further enhance the understanding and applicability of these models in real-world assembly tasks.

\section*{Acknowledgments}
This research is sponsored by national funds through FCT – Fundação para a Ciência e a Tecnologia, under the project UIDB/00285/ 2020 and LA/P/0112/2020, and the grant 2021.08012.BD.

%Bibliography
\bibliographystyle{unsrt}  
\bibliography{references}

\end{document}